    \RequirePackage{fix-cm}
    \documentclass[twocolumn]{svjour3}          % twocolumn
    \smartqed  % flush right qed marks, e.g. at end of proof
    \usepackage{setspace}
\usepackage{capt-of}% or \usepackage{caption}
\usepackage{floatrow}
\usepackage{cite}
\usepackage{natbib}
\usepackage{graphicx}
\usepackage{float}
\usepackage[normalem]{ulem}
\usepackage{subfigure}
\usepackage{amsfonts}
\usepackage{ifthen}
\usepackage{booktabs}
\usepackage{amssymb}
\usepackage{supertabular}
\usepackage{array}
\usepackage{multirow}
\usepackage{fancyhdr}
\usepackage{psfrag}
\usepackage{amsmath}
\usepackage{hhline}
\usepackage{type1cm}
\usepackage{lettrine}
\usepackage{cite}
\usepackage{fancybox}
\usepackage{enumitem}
\usepackage{bm}
\usepackage{pifont}
\usepackage{threeparttable}
\usepackage{soul}

\usepackage{tikz}
\usetikzlibrary{shapes.geometric, arrows}
\usepackage[linesnumbered,ruled,vlined]{algorithm2e}
\usepackage[noend]{algpseudocode}
\usepackage{bchart}
\usepackage{mathtools}
\usepackage{makecell}

\usepackage{pifont}

\newcommand{\ie}{\textit{i}.\textit{e}.}
\newcommand{\eg}{\textit{e}.\textit{g}.}

\usepackage{xcolor}
\usepackage{color, colortbl}
\definecolor{citecolor}{HTML}{0071bc}
\definecolor{tabhighlight}{HTML}{e5e5e5}
\usepackage[colorlinks,citecolor=citecolor]{hyperref}
\usepackage{arydshln}

\newcommand{\sysName}{AniClipart}

    \makeatletter
    \renewcommand\paragraph{
      \@startsection{paragraph} % name
      {4} % level
      {\z@} % indent
      {.5em \@plus1ex \@minus.2ex} % beforeskip
      {-.5em} % afterskip
      {\normalfont\normalsize\bfseries} % style
    }
    \makeatother

\begin{document}
    \sloppy
    \title{AniClipart: Clipart Animation with Text-to-Video Priors}
    \author{Ronghuan Wu \and
            Wanchao Su  \and
            Kede Ma     \and
            Jing Liao
    }
    \institute{Ronghuan Wu \at
                  City University of Hong Kong \\
                  \email{rh.wu@my.cityu.edu.hk}           %  \\
               \and
               Wanchao Su \at
                  Monash University \\
                  \email{wanchao.su@monash.edu}
               \and
               Kede Ma \at
                  City University of Hong Kong \\
                  \email{kede.ma@cityu.edu.hk}
               \and
               Jing Liao \at
                  City University of Hong Kong \\
                  \email{jingliao@cityu.edu.hk}
    }
    
    \date{}

    \maketitle
    
    \newcommand*{\source}{source}
    % %--------------------------------------------------------
    \begin{abstract}
Clipart, a pre-made art form, offers a convenient and efficient way of creating visual content. However, traditional workflows for animating static clipart are laborious and time-consuming, involving steps like rigging, keyframing, and inbetweening.
Recent advancements in text-to-video generation hold great potential in resolving this challenge. Nevertheless, direct application of text-to-video models often struggles to preserve the visual identity of clipart or generate cartoon-style motion, resulting in subpar animation outcomes.
In this paper, we introduce \sysName, a computational system that converts static clipart into high-quality animations guided by text-to-video priors. To generate natural, smooth, and coherent motion, we first parameterize the motion trajectories of the keypoints defined over the initial clipart image by cubic B\'{e}zier curves. We then align these motion trajectories with a given text prompt by optimizing a video Score Distillation Sampling (SDS) loss and a skeleton fidelity loss. 
By incorporating differentiable As-Rigid-As-Possible (ARAP) shape deformation and differentiable rendering, \sysName~can be end-to-end optimized while maintaining deformation rigidity. Extensive experimental results show that the proposed \sysName~consistently outperforms the competing methods, in terms of text-video alignment, visual identity preservation, and temporal consistency. Additionally, we showcase the versatility of \sysName~by adapting it to generate layered animations, which allow for topological changes.
Our code is available at \url{https://aniclipart.github.io/}.

\end{abstract}

    % %--------------------------------------------------------

    % %--------------------------------------------------------
    \begin{figure*}[t]
    \centering
    \includegraphics[width=1.0\linewidth]{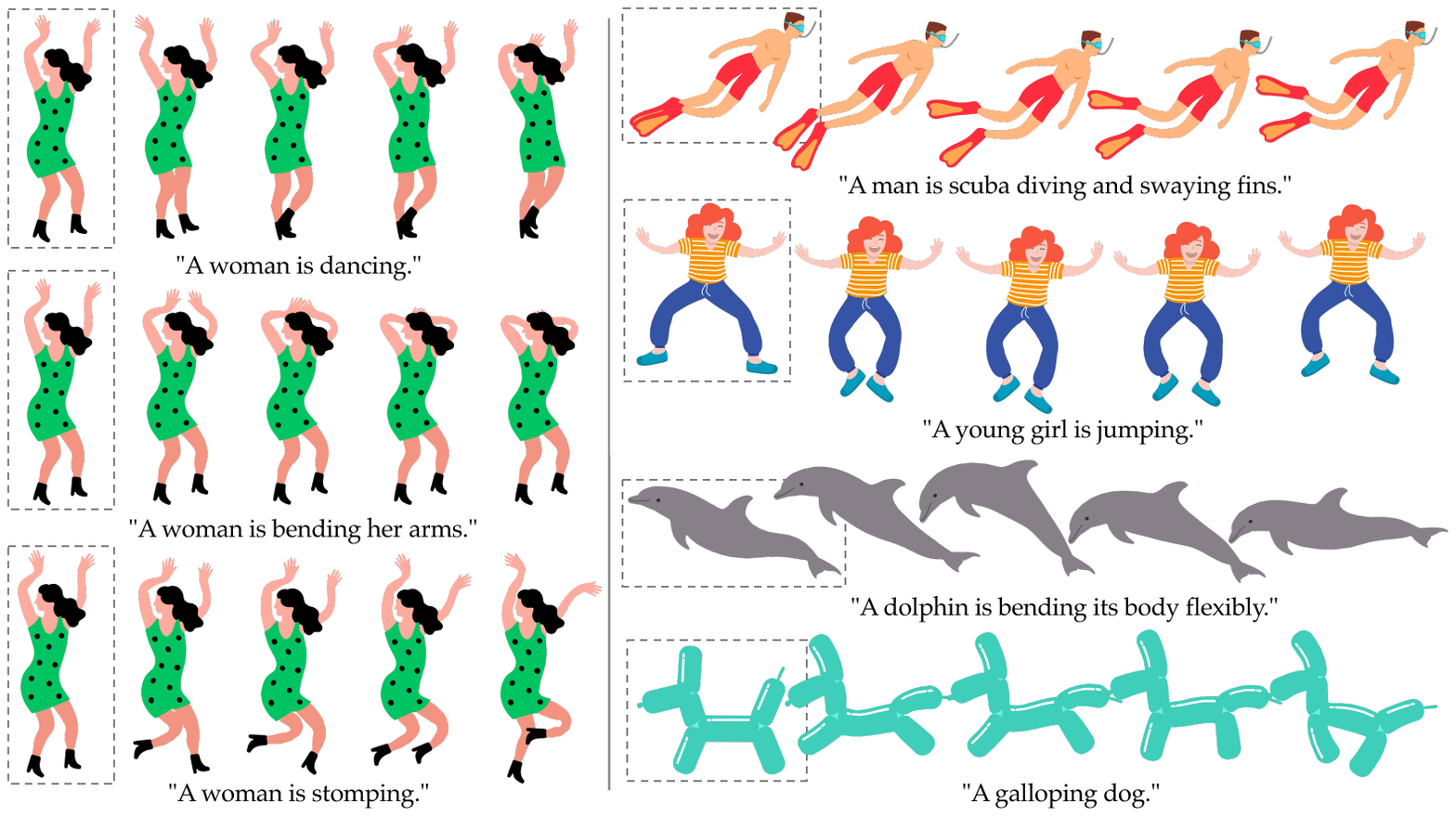}
    \caption{\sysName~creates high-quality clipart animations guided by text prompts with visual identity preservation and temporal motion consistency. The left panel displays different animations generated from the same clipart input, each guided by a different text prompt. The right panel presents animations across diverse clipart categories. Initial clipart images are marked with dashed-line boxes.
    }
    \label{fig:teaser}
\end{figure*}

\section{Introduction}

Clipart, a collection of pre-made art elements, offers an accessible and efficient solution for enhancing visual presentation without the need for customized artwork. Its simplicity, versatility, and high quality make clipart an invaluable resource for visual communication and creativity in the fields of education, business, and web design. Animated clipart takes the functionality of static clipart a step further by adding motion, making it more dynamic, engaging, enjoyable, and memorable in interactive spaces.

Animating clipart has traditionally been a meticulous process, involving steps such as rigging, keyframing, inbetweening, and precise control of spacing and timing. Recent advancements in Text-to-Image (T2I) generation, exemplified by Stable Diffusion~\citep{rombach2022high}, have revolutionized and streamlined the creation of high-quality clipart. Although proven successful in some instances, the automatic animation of clipart remains a largely untapped area of research.

The growing demand for animated clipart, coupled with its labor-intensive creation, highlights the need for a computational system that can animate static clipart with minimal to no manual intervention. Recent Text-to-Video (T2V) models~\citep{xing2023dynamicrafter, chen2023videocrafter1, chen2024videocrafter2, zhang2023i2vgen}, which take a text prompt and a bitmap image as input, present a feasible solution. However, these models are inadequate for generating high-quality clipart animations due to a substantial mismatch between statistics of natural videos and clipart animations. T2V models are predominantly trained on natural videos of high fidelity and spatiotemporal complexity. In contrast, clipart animations are cartoon-style with an emphasis on simplicity and clarity. As a consequence, T2V models often fail to maintain both the spatial visual identity and temporal consistency, generating animations with unwanted rich textures and flickering artifacts. 

To address the aforementioned challenges, we introduce \emph{\sysName}, a computational system leveraging pretrained T2V diffusion models to animate clipart and align it with given text prompts. Inspired by the standard animation pipeline, the key to the success of \sysName~is to designate keypoints on the initial clipart image, and predict their trajectories with the help of T2V models for animation. 
Specifically, we parameterize keypoint motion trajectories by cubic B\'{e}zier curves to enable smooth motion prediction while maintaining manageable complexity.
To align motion with a given text prompt, we rely on a video Score Distillation Sampling (SDS) loss to optimize the parameters of the B\'{e}zier trajectories (\ie, 2D coordinates of control points), exploiting implicit motion priors in pretrained T2V diffusion models. 
To encourage coherent motion synthesis for all keypoints, we incorporate a skeleton fidelity loss in addition to the video SDS loss. This loss penalizes large bone length variations in the skeleton formed by the keypoints. Additionally, we integrate differentiable As-Rigid-As-Possible (ARAP) shape deformation~\citep{Takeo2005arap} and differentiable rendering~\citep{li2020DiffVG} to facilitate end-to-end optimization of \sysName. The use of ARAP ensures rigid warping of clipart to adapt to new poses, while differentiable rendering is necessary for vector clipart to allow gradient backpropagation through the non-differentiable rasterization step, compatible with the video SDS loss that operates on bitmap inputs.

Extensive experiments and ablation studies demonstrate \sysName's ability to generate vibrant, engaging, and cartoon-like clipart animations across a wide range of visual content, including humans, animals, and objects (see  Fig.~\ref{fig:teaser}).
\sysName~also supports layered animation to handle movements involving topological changes. In summary, our contributions can be summarized as follows.
\begin{enumerate}
    \item 
    We introduce \sysName, a computational system capable of generating high-quality clipart animations based on text prompts, marking a step forward in automated animation generation.
    \item 
    We successfully harness the implicit motion priors of T2V diffusion models through the video SDS loss, complemented by the skeleton fidelity loss, to produce text-aligned, coherent motion in abstract, cartoon-like style.
    \item
    We integrate differentiable ARAP shape deformation and differentiable rendering to enable end-to-end optimization of \sysName.
\end{enumerate}
    % %--------------------------------------------------------

    %--------------------------------------------------------
    \section{Related Work}
\label{sec:related_work}

In this section, we provide an overview of prior work that is closely related to ours: 2D animation (Sec. \ref{related_2d_anim}), T2V generation (Sec. \ref{related_t2v}), and SDS-driven applications (Sec. \ref{related_sds}).

\subsection{2D Animation}
\label{related_2d_anim}

\begin{figure}[t]
    \centering
    \includegraphics[width=1.0\linewidth]{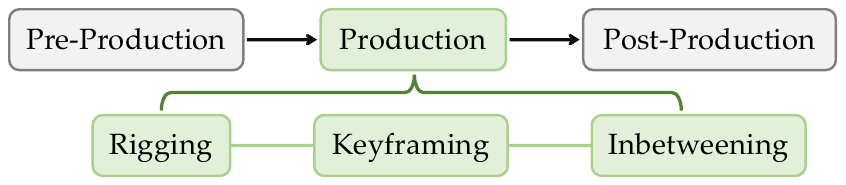}
    \caption{
    A simplified animation production pipeline.
    }
    \label{fig:anim_pipeline}
\end{figure}

Traditional animation production is cost-intensive and time-consuming.
Fig.~\ref{fig:anim_pipeline} illustrates a simplified pipeline for 2D animation, emphasizing three crucial steps: rigging, keyframing, and inbetweening. Recent research has focused on developing computational algorithms to automate these steps.

\textit{Rigging}
is an initial step in animation that constructs object skeletons.
Along with skinning algorithms~\citep{forstmann2006fast, kavan2007skinning, le2019direct}, motion applied to the skeleton can propagate seamlessly across the entire object. Automated rigging techniques can be broadly classified into two categories: template-based methods and template-free methods.
Template-based approaches map a predefined skeletal template onto an object~\citep{baran2007automatic, li2021learning}. While effective, they lack adaptability for objects that deviate significantly from the template.
Template-free approaches can generate rigs for arbitrary objects. Early approaches~\citep{au2008skeleton, huang2013l1, tagliasacchi2012mean} analyze mesh features to produce curve-skeletons, but they often fail to account for movable parts.
Recently, deep learning-based approaches~\citep{liu2019neuralskinning, xu2020rignet} exhibit strong potential for rig generation, offering a powerful alternative for animators.

\textit{Keyframing} involves specifying object poses at keyframes in animation. 
Animators first create a deformable puppet, typically represented by a triangle mesh, and manually manipulate handles (\ie, control points generated in rigging) across the keyframes.
However, defining desired poses remains a challenge for individuals without animation expertise~\citep{fan2018tooncap}. To address this, methods that transfer motion onto predefined puppets have been proposed.~\citet{bregler2002turning} and~\citet{de2006re} explored techniques for capturing and transferring motion from existing cartoons.~\citet{hornung2007character} animated 2D characters using 3D motion capture data.
Animated Drawings~\citep{smith2023method} creates rigged characters from children's artwork, and animates them using predefined human motion.
Recent innovations have also focused on extracting motion from videos. Live Sketch~\citep{su2018live} maps video-derived motion onto sketches,
Pose2Pose~\citep{willett2020pose2pose} applies clustering to select key poses from performance videos,
and AnaMoDiff~\citep{tanveer2024anamodiff} warps objects based on optical flow computed from a driving video.

\textit{Inbetweening} refers to inserting drawings between keyframes, transforming choppy, disjointed animations into smooth and fluid ones. 
A common approach involves shape interpolation between keyframes~\citep{Alexa2000AsrigidaspossibleSI, baxter2008rigid, baxter2009n, whited2010betweenit, fukusato2022view, fukusato2016active, kaji2012mathematical, chen2013planar}.
As a special case of frame interpolation, recent advances in video frame interpolation~\citep{liu2017video, niklaus2017video1, niklaus2017video2, jiang2018super, niklaus2018context, xu2019quadratic, niklaus2020softmax, park2020bmbc, sim2021xvfi, huang2022real, reda2022film, lu2022video} can be directly transferred to animation inbetweening. However,~\citet{siyao2021deep} highlighted the distinct features of animation videos, such as completely flat color regions and exaggerated motion, which often cause natural video interpolation methods to underperform, and call for animation-specific
algorithms~\citep{siyao2021deep, li2021deep, chen2022improving, xing2024tooncrafter}.

In animation production, objects can be depicted using either bitmap images or vector graphics. While both formats share a similar animation pipeline, vector graphics animation requires extra effort, with the development of improved data structures~\citep{dalstein2015vector}, efficient representation learning algorithms~\citep{carlier2020deepsvg, cao2023svgformer}, specialized skinning techniques~\citep{liu2014skinning}, and advanced inbetweening methods~\citep{siyao2023deep}. In this paper, we propose \sysName~that automates rigging, keyframing, and inbetweening, and supports clipart animation for both bitmap and vector objects.

\subsection{T2V Generation}
\label{related_t2v}
Generating videos from text descriptions is challenging due to the inherent ambiguity and multimodal complexity of the task, the scarcity of high-quality text-video pairs, and the high computational cost involved. 
The progress of T2V models mirrors that of T2I approaches, ranging from autoregressive generation~\citep{villegas2022phenaki, wu2022nuwa, hong2022cogvideo, kondratyuk2023videopoet} to denoising diffusion~\citep{ho2022imagen, gupta2023photorealistic}.
A straightforward idea is to learn a text-conditioned probability model over the video data in raw pixel or latent domain~\citep{singer2022make, ho2022imagen}. However, this entails considerable training time and computational expense. A more computationally efficient alternative is to integrate additional temporal modules, such as temporal adapters~\citep{guo2023animatediff}, pseudo-3D convolutions~\citep{singer2022make}, and temporal attention mechanisms~\citep{ho2022imagen, singer2022make, guo2023animatediff, girdhar2023emu, blattmann2023align} into pretrained T2I models~\citep{rombach2022high}. The added modules can be trained solely using text-video data pairs or along with the T2I model~\citep{singer2022make, blattmann2023align, girdhar2023emu, ge2023preserve, yuan2024inflation, guo2023animatediff}.
As a practical extension, certain methods accept an additional image as input to enable image-conditioned T2V generation~\citep{blattmann2023stable, gu2023seer, wang2023videocomposer, xing2023dynamicrafter, zhang2023i2vgen}.
This is typically achieved by directly concatenating the image with the initial noise~\citep{xing2023dynamicrafter} or employing cross-attention~\citep{guo2023sparsectrl} in a similar spirit to ControlNet~\citep{zhang2023adding}.

Despite astonishing results for certain text prompts, current T2V models show limited generalizability in terms of text-video alignment and perceived video quality. These limitations become more pronounced when the text prompt is highly detailed and precise, or the input image contains rich structures and textures. Direct application of these models for clipart animation adds additional complexity because of the mismatch in statistics between natural videos and clipart animations.
Most recently,~\citet{li2024generative} focused on synthesizing simple natural oscillatory motion (\eg, objects swaying in the wind) using a diffusion model in the frequency domain. Nevertheless, this method is less suited for non-oscillatory, cartoon-like, and semantically meaningful motion we are looking for.
In this paper, we propose to exploit the motion priors embedded in pretrained T2V models for optimization of the keypoint motion trajectories. Together with the skeleton fidelity loss, the optimized \sysName~enables high-quality, text-conditioned clipart animation.

\begin{figure*}
    \centering
    \includegraphics[width=0.90\linewidth]{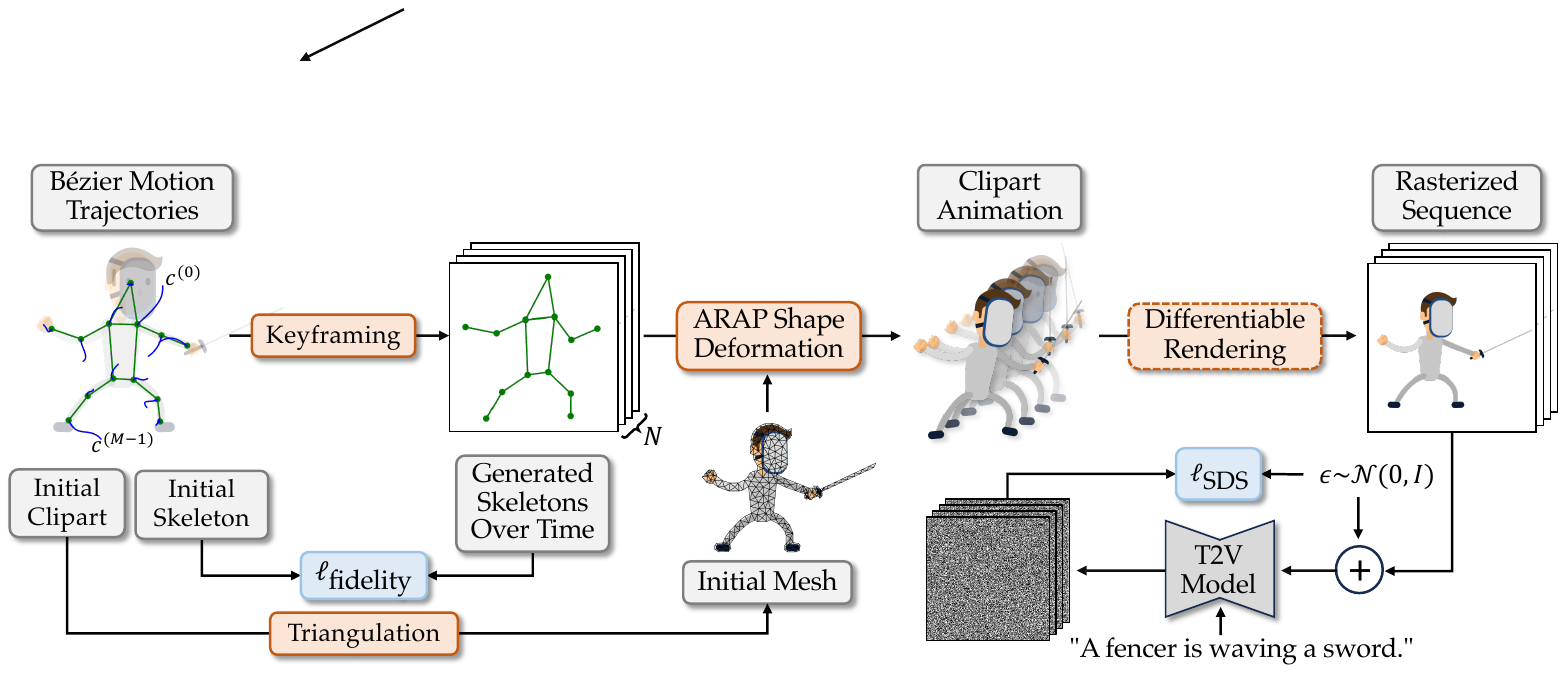}
    \caption{
    \textbf{System Diagram of \sysName}.
    Given an initial clipart image with $M$ keypoints, we initialize $M$ corresponding cubic B\'{e}zier motion trajectories, parameterized by  $\{c^{(i)}\}_{i=0}^{M-1}$.
    For a sequence of $N$ frames, keypoints are updated at each frame by sampling along these trajectories. The displaced keypoints are responsible for driving the ARAP shape deformation algorithm, which warps the object, represented by a triangle mesh, into new poses.
    This gives rise to a clipart animation, which is (optionally rasterized and) passed to a T2V model to compute the video SDS loss.
    To ensure motion coherence across all keypoints, a skeleton fidelity loss is also applied, penalizing changes in bone lengths over time.
    }
    \label{fig:pipeline}
\end{figure*}

\subsection{SDS-Driven Applications}
\label{related_sds}
T2I diffusion models, pretrained on vast text-image pairs, enable 2D image synthesis from text prompts. Their use now extends to other synthesis tasks, particularly those with limited paired data.
DreamFusion~\citep{poole2022dreamfusion} generates 3D NeRF-like representations from text descriptions by optimizing an image SDS loss. It is a simplified variant of the diffusion model training loss, where the gradient is taken with respect to the input image, excluding the U-Net Jacobian term for computational efficiency.
Since its inception, the image SDS loss has been widely used for various generation tasks, including artistic typography~\citep{iluz2023word, tanveer2023ds}, vector graphics~\citep{jain2022vectorfusion}, sketches~\citep{xing2023diffsketcher, qu2023sketchdreamer}, meshes~\citep{chen2023fantasia3d}, and texture maps~\citep{metzer2023latent, tsalicoglou2023textmesh}.
The advent of T2V diffusion models~\citep{ni2023conditional, wang2023videocomposer, dai2023fine, chen2023videocrafter1, chen2024videocrafter2, bar2024lumiere} has naturally broadened the application scope of the SDS loss to the video domain.
For example, it has been applied to create vector sketch animations~\citep{gal2023breathing}.
However, this technique is less effective for clipart animation due to its inability to maintain shape rigidity. In this paper, we successfully apply the video SDS loss for text-driven clipart animation.

    %--------------------------------------------------------

    %--------------------------------------------------------
    \section{\sysName}

In this section, we introduce our clipart animation system, \sysName. We begin with a method overview (Sec. \ref{method_overview}), followed by a detailed description of clipart preprocessing (Sec. \ref{method_data_preperation}), B\'{e}zier-parameterized animation (Sec. \ref{method_bezier}), and the loss functions used (Sec. \ref{method_loss}).

\subsection{Method Overview}
\label{method_overview}
In clipart preprocessing, we detect keypoints, build a skeleton, and construct a triangle mesh over the initial clipart image. We then parameterize the keypoint motion trajectories using cubic B\'{e}zier curves.
When the keypoints move to new positions at specific time instances, the differentiable ARAP shape deformation algorithm~\citep{Takeo2005arap} is employed to adjust the entire object driven by the displaced keypoints. By sampling along the continuous B\'{e}zier trajectories, we obtain a clipart animation, which is sent to a pretrained T2V model~\citep{wang2023modelscope} to compute the video SDS loss~\citep{poole2022dreamfusion}.
Along with a skeleton fidelity loss that encourages coherent motion across all keypoints, we optimize the parameters of B\'{e}zier trajectories (\ie, 2D coordinates of control points) for clipart animation (see Fig.~\ref{fig:pipeline}).

\begin{figure}[t]
    \centering
    \includegraphics[width=1.0\linewidth]{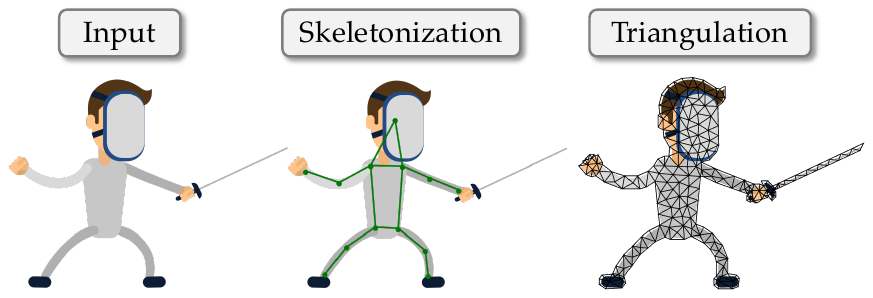}
    \caption{Template-based and anatomically meaningful keypoint detection by UniPose~\citep{yang2023unipose} for articulated objects (\eg, humans), followed by skeletonization and triangulation.
    }
    \label{fig:data_prep}
\end{figure}

\subsection{Clipart Preprocessing}
\label{method_data_preperation}
Akin to traditional animation production, the initial step in our clipart animation pipeline involves object rigging, in which we detect keypoints on clipart and construct a skeleton between these points. 
Existing keypoint detection algorithms~\citep{sun2023uniap, ye2022superanimal, mathis2018deeplabcut, ng2022animal, xu2022vitpose, jiang2023rtmpose, yang2023boosting} excel at assigning template keypoints to articulated characters, but are limited to object categories within their training datasets. This poses a challenge when applied to clipart, which encompasses a wide range of cartoon objects.

\begin{figure}[t]
    \centering
    \includegraphics[width=1.0\linewidth]{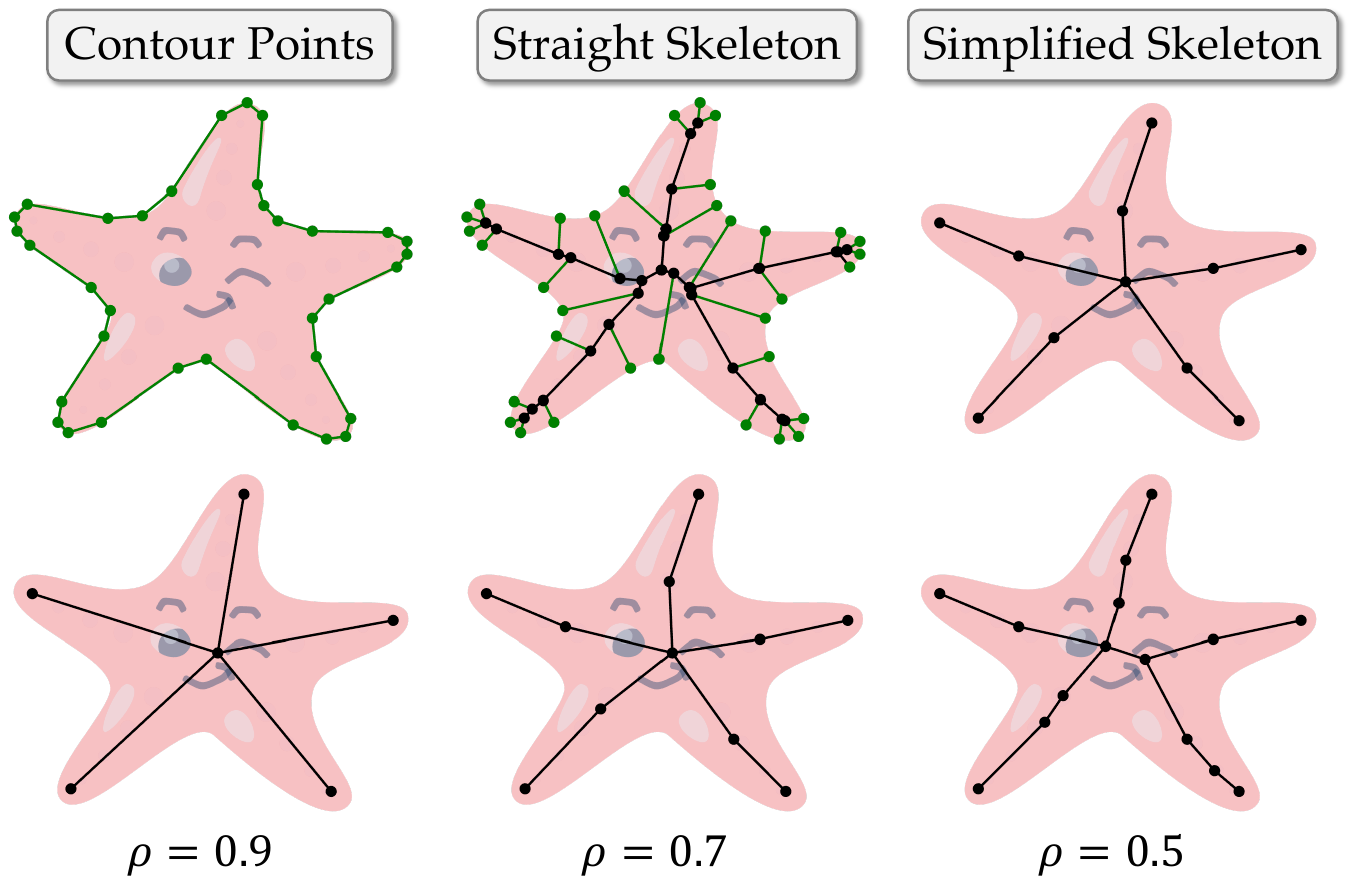}
    \caption{
    Template-free keypoint detection and skeleton construction. The first row shows a visual example of an invertebrate starfish, while the second row highlights the impact of different $\rho$ values.
    }
    \label{fig:keypoint}
\end{figure}

Here, we adopt a hybrid approach to combine the strengths of existing keypoint detection methods. Specifically, we leverage the template-based UniPose algorithm~\citep{yang2023unipose} to detect keypoints and construct skeletons for articulated objects (\eg, humans and quadrupedal animals). The identified keypoints often have clear anatomical meanings like joints and limb endpoints (see Fig.~\ref{fig:data_prep}). For broader object categories, like sea animals and plants, we employ an alternative keypoint detection algorithm, which involves three steps (see Fig.~\ref{fig:keypoint}).
\begin{enumerate}
    \item \textit{Binarization and Boundary Detection}. We binarize color clipart, where the objects are displayed in black and the background in white. Using the \texttt{findContours()} function\footnote{\url{https://docs.opencv.org/3.4/df/d0d/tutorial_find_contours.html}} in OpenCV, we detect points along the boundaries of the objects, which are then connected to form the contour edges.
    \item \textit{Skeleton Generation}. We generate a straight skeleton by propagating the contour edges inward in their perpendicular directions. During this process, all edges move at the same constant speed, and the collapsed edges form the keypoints~\citep{cacciola2004cgal}, denoted as $\{p_0^{(i)}\}_{i=0}^{M-1}$, where $M$ is the total number of keypoints. Here, we use the \texttt{scikit-geometry} library for implementation.
    \item \textit{Skeleton Simplification}. The initial skeleton often contains excessive bones (\ie, edges between keypoints), which are less suitable for animation purposes. To address this, we prune unnecessary outer bones (highlighted by the green line segments in Fig.~\ref{fig:keypoint}) and simplify inner bones using edge collapsing, which iteratively merges two adjacent keypoints if their distance is below a predetermined threshold. Typically, we set the threshold to be proportional to the average bone length:
    \begin{align}\label{eq:thre}
        \delta  =\frac{\rho}{\vert\mathcal{E}\vert}\sum_{(i,j)\in\mathcal{E}}\left\Vert p^{(i)}_0 - p^{(j)}_0\right\Vert_2,
    \end{align}
    where $p^{(i)}_0$ and $p^{(j)}_0$ represent a pair of adjacent keypoints forming a bone in the skeleton, whose indices are stored in the set $\mathcal{E}$ with a cardinality of  $\vert\mathcal{E}\vert$. $\rho$ is a linear scaling factor. Skeleton simplification proceeds in multiple iterations until no further merging is possible.
    Adjusting $\rho$ allows for the generation of skeletons with varying levels of complexity (see the second row of Fig.~\ref{fig:keypoint}). 
\end{enumerate}

After detecting keypoints and constructing the skeleton, we apply a triangulation algorithm~\citep{shewchuk1996triangle} to generate a triangle mesh for the initial object, which completes clipart preprocessing (see Fig.~\ref{fig:data_prep}).

\begin{figure}[t]
    \centering
    \includegraphics[width=0.9\linewidth]{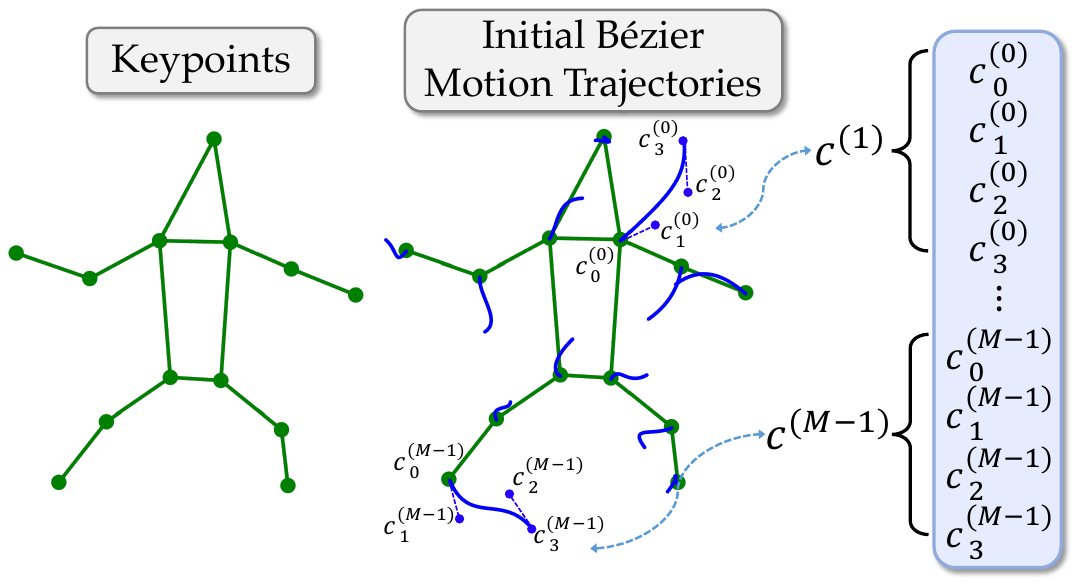}
    \caption{\textbf{B\'{e}zier Trajectory Initialization}.
    We parameterize each keypoint motion trajectory by a cubic B\'{e}zier curve, where the first control point (\ie, the starting point of the curve) is initialized to be the corresponding keypoint. 
    This ensures that the animation begins with the initial pose of the object. Each of the remaining three control points is sequentially initialized using a Gaussian distribution with the mean centered at the preceding control point and a small variance to induce mild motion.
    The initial trajectories are relatively short compared to the skeleton, but are intentionally amplified in the figure for improved visualization.
    }
    \label{fig:motion_path}
\end{figure}

\subsection{B\'{e}zier-Parameterized Animation} \label{method_bezier}
Prior research predicts future positions of B\'{e}zier control points that define a sketch at discrete timesteps~\citep{gal2023breathing}, which often face challenges in generating identity-preserving objects with smooth motion. In stark contrast, we resort to parametric functions that are continuous and differentiable to model the motion trajectories of much fewer keypoints rather than all control points, which better regularizes and substantially simplifies this temporal prediction task.
Here, we choose to parametrize keypoint motion trajectories as cubic B\'{e}zier curves due to their simplicity (requiring only a small set of control points), flexibility to model both simple and complex motion by adjusting the number of control points, smooth interpolation between keyframes, and their widespread use and ease of implementation in design applications.
Specifically, we assign a cubic B\'{e}zier trajectory to each keypoint, and align it to the starting control point, ensuring that the animation begins with the initial object pose.
Each of the remaining three control points is randomly initialized using a Gaussian distribution with the mean centered at the preceding control point and a small variance to introduce moderate motion (see Fig.~\ref{fig:motion_path}).

Formally, the initial clipart image $x_0$ contains $M$ keypoints, $\{p_0^{(i)}\}_{i=0}^{M-1}$, accompanied by $M$ cubic B\'{e}zier trajectories, parameterized by $\{c^{(i)}\}_{i=0}^{M-1}$, each of which is controlled by four points $\{c_j^{(i)}\}_{j=0}^3$.
To index keypoint motion over time, we define a sequence of timesteps, $\{1,2,\dots,N-1\}$, where $N$ denotes the total number of frames in the animation.
At the $t$-th timestep, we sample points along the B\'{e}zier trajectories as the predictions of the future keypoints, denoted as  $\{p_t^{(i)}\}_{i=0}^{M-1}$.
These updated keypoints drive the animation of the clipart object, modeled as a triangle mesh, using the ARAP shape deformation algorithm~\citep{Takeo2005arap}.
ARAP provides an intuitive approach to deforming 2D shapes by either interactively dragging keypoints or automatically positioning them (as demonstrated in \sysName). ARAP operates through two computationally efficient steps, both of which admit closed-form solutions: it first computes the translation and rotation for each triangle, and then adjusts the scale to minimize geometric distortions. Importantly, ARAP is inherently differentiable, and we implement it with automatic differentiation to enable seamless gradient backpropagation. After performing ARAP shape deformation at the $t$-th timestep, we warp the initial clipart image according to the updated triangle mesh using either linear texture mapping (for bitmap images) or precomputed barycentric coordinates (for vector graphics). This generates the $t$-th clipart frame, denoted as $x_t$, in the final animation $x$.

In practice, clipart animations often feature looping designs due to their reusability (allowing indefinite replay), visual appeal (with natural and rhythmic motion), and efficiency in file size optimization  (particularly for web-based and resource-constrained applications). The proposed \sysName~is well-suited for generating looping animations by mirroring the first half of the animation in reverse-time order during optimization. When looping is not required, \sysName~is optimized to directly generate a clipart animation $x$ of $N$ frames. Last, if $x$ is in the format of vector graphics,  differentiable rendering~\citep{li2020DiffVG} is applied to convert it into a bitmap video, ensuring compatibility with the video SDS loss, detailed subsequently.

\subsection{Loss Functions}
\label{method_loss}
We utilize the video SDS loss to drive text-conditioned clipart animation, while encouraging motion coherence of all keypoints through a skeleton fidelity loss.

The \textbf{Video SDS Loss} is built upon the weighted denoising score matching objective, with the input video $x$ as parameters: 
\begin{equation}\label{eq:lSDS}
    \ell_{\mathrm{SDS}}(x) = \mathbb{E}_{t, \epsilon} \left[w(t)\left\Vert \epsilon_\phi(\alpha_t x+ \beta_t \epsilon;y,t) - \epsilon  \right\Vert_2^2 \right],
\end{equation}
where $t$ is sampled from a discrete uniform distribution $\mathcal{U}\{0,T\}$ and $T$ is the number of forward diffusion steps. $\epsilon$ is a standard Gaussian noise vector sampled from $\mathcal{N}(0,I)$. $w(t)$ is a weighting function, depending on the timestep $t$.  $\epsilon_\phi(\cdot)$ is the U-Net denoising network in the T2V model~\citep{wang2023modelscope}, conditioned on the text prompt $y$ and the timestep $t$.
$\{\alpha_t\}_{t =1}^T$ represent a monotonically increasing noise schedule, where $\alpha_t^2 + \beta_t^2 = 1$, $\alpha_0 \approx 1$, and $\alpha_T \approx 0$.
At a high level, the SDS loss leverages (the gradient of) a diffusion model's score function to guide the optimization of a different model (in our case, \sysName) toward generating 
outputs that align with the desired text-conditioned distribution. The video SDS loss extends the image SDS loss to the video domain, leveraging motion information to ensure temporal consistency in video generation. Optimization of \sysName~entails differentiating the loss in Eq.~\eqref{eq:lSDS} with respect to the output of the U-Net, $\epsilon_\phi(\cdot)$, and propagating the gradient through $\epsilon_\phi(\cdot)$ to the output of \sysName, $x$, and finally to the parameters of \sysName, which are collectively denoted by $c = \{\{c^{(i)}_j\}_{j=0}^3\}_{i=0}^{M-1}$. This extended chain of computation not only demands substantial computational resources, but also becomes unstable when the U-Net Jacobian is poorly conditioned at small noise levels. To resolve this, \citet{poole2022dreamfusion} proposed a gradient approximation by omitting the U-Net Jacobian:
\begin{equation}\label{eq:sdsg}
    \nabla_c \ell_{\mathrm{SDS}}(x(c)) = \mathbb{E}_{t, \epsilon} \left[w(t)(\epsilon_\phi(\alpha_t x +\beta_t \epsilon;y,t) - \epsilon) \frac{\partial x}{\partial c} \right],
\end{equation}
where the dependence of the output clipart animation $x$ of \sysName~on its parameters $c$ is made explicit.
As a practical trick, $\epsilon_\phi(\alpha_t x +\beta_t \epsilon;y,t)$ in Eq.~\eqref{eq:sdsg} is further modified using classifier-free guidance~\citep{ho2022classifier} for better text-video alignment:
\begin{align}\label{eq:cfg}
    \epsilon_\phi(\alpha_t x+ \beta_t \epsilon;y,t) &\leftarrow(1+s)\epsilon_\phi(\alpha_t x+ \beta_t \epsilon;y,t) \notag \\
    &\quad - s\epsilon_\phi(\alpha_t x+ \beta_t \epsilon;\emptyset, t).
\end{align}
where $s$ is the classifier-free guidance parameter and $\emptyset$ denotes a null text prompt (\ie, without conditioning on any text input).

\noindent{\textbf{Skeleton Fidelity Loss}}.
Optimizing keypoint motion trajectories with the video SDS loss $\ell_{\mathrm{SDS}}$ produces clipart animations well aligned with input text prompts. 
Nevertheless, these animations occasionally exhibit local geometric distortions (\eg, the overly retracted neck in Fig.~\ref{fig:ablation}), as different keypoints may be optimized to move incoherently. 
To better preserve object fidelity, we leverage the constructed skeleton (Sec. \ref{method_data_preperation}), and penalize deviations in bone lengths from the initial configuration:
\begin{align}\label{eq:fidelity}
    \ell_{\mathrm{fidelity}}(p) =& \frac{1}{(N-1)\vert\mathcal{E}\vert } 
    \sum_{t=1}^{N-1} \sum_{(i,j)\in \mathcal{E}} \bigg( 
    \left\Vert p^{(i)}_t - p^{(j)}_t\right\Vert_2 \notag \\
    &\quad - \left\Vert p_0^{(i)} - p_0^{(j)}\right\Vert_2 \bigg)^2,
\end{align}
where $p^{(i)}_t$ and $p^{(j)}_t$ are a pair of adjacent keypoints forming a bone at the $t$-th frame. Given our choice of motion parameterization,  $p^{(i)}_t$ can be efficiently computed using the cubic B\'{e}zier control points:
\begin{align}
    p^{(i)}_t(c) &= (1-t/N)^3 c_0^{(i)} + 3(1-t/N)^2 (t/N) c_1^{(i)} \notag \\
    &\quad + 3(1-t/N) (t/N)^2 c_2^{(i)} + (t/N)^3 c_3^{(i)},
\end{align}
and $p^{(j)}_t$ can be computed accordingly.

The {\textbf{Overall Loss}} for \sysName~is defined as a weighted linear sum of the video SDS loss and the skeleton fidelity loss:
\begin{equation}\label{eq:loss}
    \ell_{\mathrm{total}}(c) = \ell_{\mathrm{SDS}}(x(c)) + \lambda\ell_{\mathrm{fidelity}}(p(c)),
\end{equation}
where the weighting factor $\lambda$ is set to strike a balance between the magnitudes of the two terms. \sysName~is fully differentiable, allowing end-to-end optimization of the B\'{e}zier trajectory parameters for clipart animation.

    %--------------------------------------------------------

    % %--------------------------------------------------------
    \section{Experiments}
We conduct extensive experiments to evaluate the effectiveness of our proposed \emph{\sysName}. We first introduce the experimental setups (Sec.~\ref{experiment_training}) and the evaluation metrics (Sec.~\ref{metrics}). Next, we compare \sysName~with a closely related method for sketch animation~\citep{gal2023breathing}, as well as state-of-the-art T2V models (Sec.~\ref{experiment_comparison}). We then conduct a series of ablation studies to justify the design choices of \sysName~(Sec.~\ref{experiment_ablation}), and demonstrate its ability to handle more challenging animation cases with topological changes (Sec.~\ref{extension}).

\subsection{Experimental Setups}
\label{experiment_training}

We collected $30$ clipart illustrations from Freepik\footnote{\url{https://www.freepik.com/}}, including $10$ humans, $10$ animals, and $10$ objects, each resized to $256\times 256$. The linear scaling factor $\rho$ in Eq.~\eqref{eq:thre} is set to $0.7$. 
The implementation of the video SDS loss relies on the ModelScope T2V model~\citep{wang2023modelscope}, where the classifier-free guidance parameter $s$ in Eq.~\eqref{eq:cfg} is set to $50$. The trade-off parameter $\lambda$ in Eq.~\eqref{eq:loss} is set to $25$. The cubic B\'{e}zier  control points $c = \{\{c^{(i)}_j\}_{j=0}^3\}_{i=0}^{M-1}$, where $M$ varies from $10$ to $13$, were optimized using Adam~\citep{kingma2014adam} over $500$ steps with a learning rate of $0.5$.
To generate clipart animations, we sampled uniformly along the optimized  B\'{e}zier motion trajectories, setting the number of frames $N$ to $24$. Animating a single clipart image on an NVIDIA RTX A6000 took approximately $25$ minutes, with a memory usage of $26$ GB.

Clipart is commonly stored and distributed in two formats: bitmap images and vector graphics. In our experiments, we focused on clipart in the Scalable Vector Graphics (SVG) format, whose scalability enables the generation of high-resolution visualization, while the layered representation facilitates the creation of more complex animations with topological changes.
Each SVG file includes multiple geometric shapes known as \textit{paths}, which are defined by sequences of control points.
During animation, we treat the control points of SVG  clipart as the barycentric coordinates within specific triangles in the mesh (see Fig.~\ref{fig:anim_pipeline}), which are moved in accordance with the corresponding warped triangles using ARAP shape deformation~\citep{Takeo2005arap}. We then employ DiffVG~\citep{li2020DiffVG} to convert SVG clipart to bitmap representation for compatibility with the video SDS loss. \sysName~can be applied straightforwardly to bitmap clipart with comparable quality. The key difference lies in the warping of bitmap frames, where all pixels (rather than the control point) in each triangle are warped to new positions, eliminating the need for a differentiable renderer.

\begin{figure*}[t]
    \centering
    \includegraphics[width=1\linewidth]{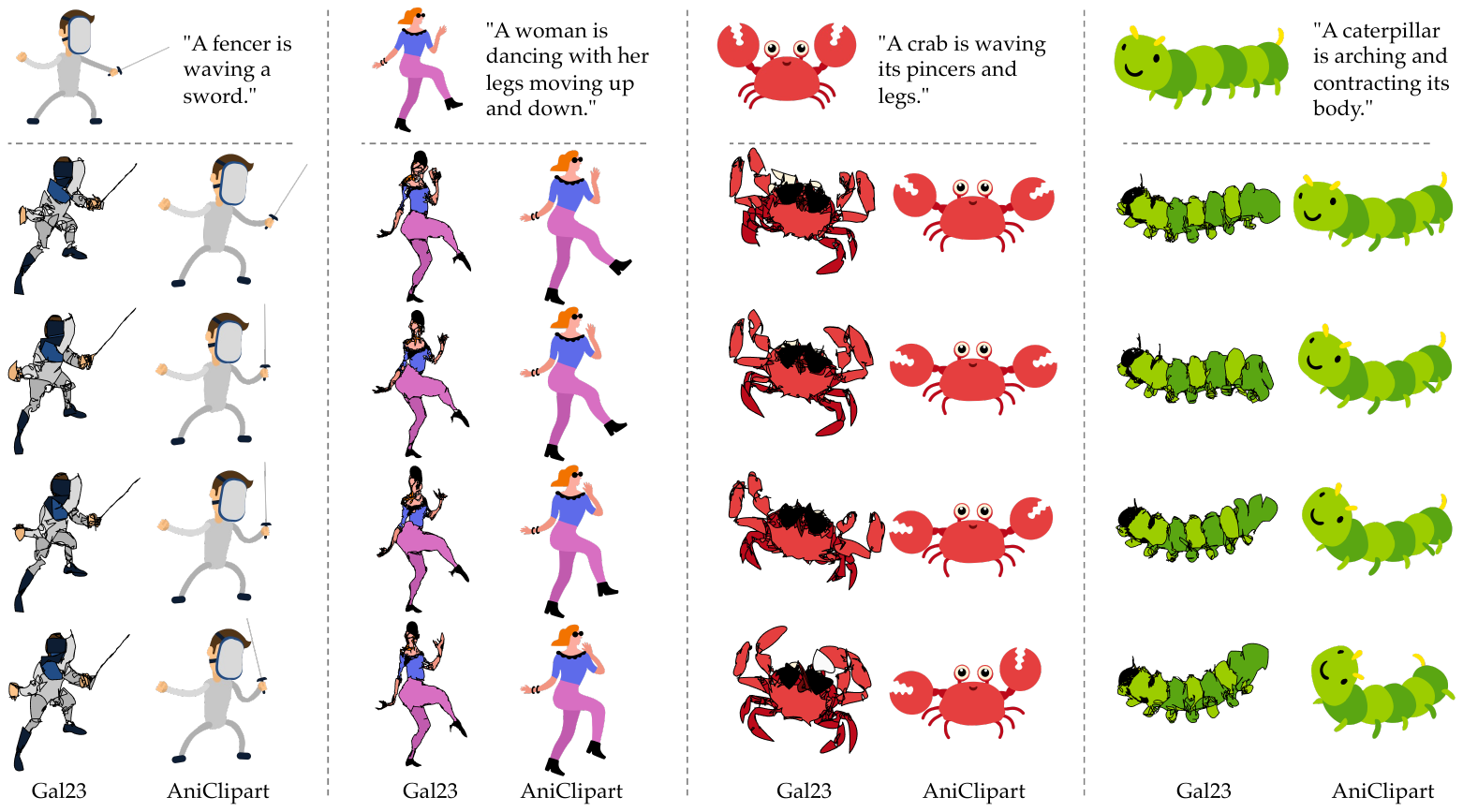}
    \caption{
    \textbf{\sysName~versus Gal23}.
    We sample four consecutive frames for visualization. Gal23 drastically distorts the overall appearance, and exhibits a lack of continuity and consistency across frames. In contrast, \sysName~effectively maintains the visual identity of the objects by preserving their overall shapes with rigidity, resulting in high-quality, text-aligned, and cartoon-like animations.}
    \label{fig:result_anisketch}
\end{figure*}

\subsection{Evaluation Metrics}
\label{metrics}

\noindent{\textbf{Bitmap Metrics}}.
We focus on evaluating two key aspects of the generated clipart animations:
(1) visual identity preservation, ensuring the generated animations maintain the visual characteristics of input clipart images; and (2) text-video alignment, ensuring the animations accurately reflect the provided text descriptions. For visual identity preservation, we employ the ViT-B/32 image encoder of the CLIP model~\citep{radford2021learning} to compute the average cosine similarity score between the feature representations of the initial frame and each generated frame in the animation. For text-video alignment, we utilize X-CLIP~\citep{ni2022expanding}, an extension of the CLIP model for the video domain, to compute the cosine similarity between text and animation feature representations.

\begin{figure*}[t]
    \centering
    \includegraphics[width=1.0\linewidth]{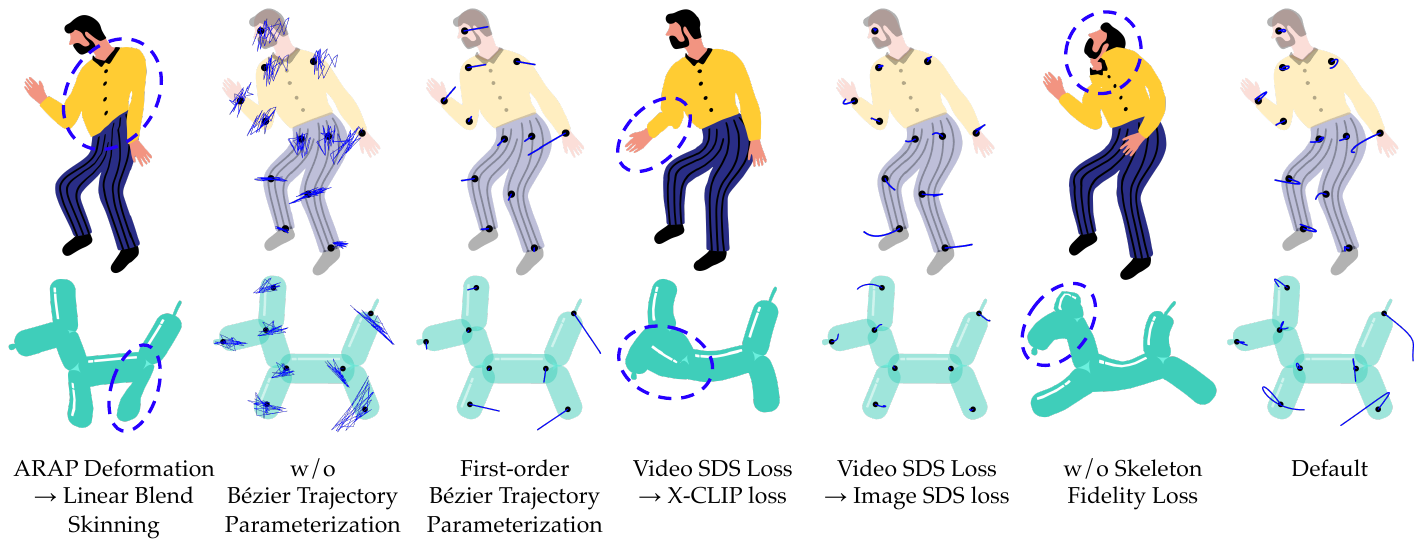}
    \caption{\textbf{Ablation Results of \sysName}.
    Replacing, simplifying, or removing key components from \sysName~can lead to animations with limited motion (\eg, the third and fifth columns), shape distortions (\eg, the first, fourth, and sixth columns), and motion inconsistencies (\eg, the second column).
    To emphasize the motion details depicted by the blue trajectories, we overlay them with the corresponding clipart image, made transparent. Shape distortions are marked using dashed-line ellipses.
    }
    \label{fig:ablation}
\end{figure*}

\noindent{\textbf{Vector Metrics}}.
Consistent with prior research~\citep{gal2023breathing}, we find that CLIP and X-CLIP fail to capture subtle geometric differences in animations. Consequently, we also incorporate three vector metrics as a more precise means of geometric assessment.
\begin{enumerate}
    \item \textit{Motion Vibrancy} measures the mean motion magnitude in an animation $x$ by computing the average length (\ie, geodesic distance) of B\'{e}zier motion trajectories:
    \begin{align}
        \mathrm{MV}(x) = \frac{1}{M}\sum_{i=0}^{M-1} \mathrm{length}\left(c^{(i)}\right).
    \end{align}
    When keypoints are directly predicted for each frame (as shown in the ``w/o B\'{e}zier Trajectory Parameterization'' column of Fig.~\ref{fig:ablation}), we connect temporally consecutive keypoints to form pseudo-trajectories, and compute the total length as the sum of the lengths of all line segments.
    Higher motion vibrancy values signify more vivid motion.
    \item \textit{Temporal Consistency}.
    The shape of vector clipart is defined by a set of control points distributed along its contours. Thus, we quantify how these control points change over time as a way of measuring temporal consistency of an animation. Denoting ${p}_t$ and ${p}_{t+1}$ as the sequences of control points at the $t$-th and $t+1$-th frames, respectively, we define the frame-wise temporal consistency by the Hausdorff distance between the control points in ${p}_t$ and ${p}_{t+1}$:
    \begin{align}
        d_H({p}_t, {p}_{t+1}) = \max\Big\{&\max_{p^{(i)}_t \in{p}_t}d_H\left(p^{(i)}_t,{p}_{t+1}\right), \nonumber\\
        &\max_{p^{(j)}_{t+1}\in{p}_{t+1}}d_H\left( p^{(j)}_{t+1},{p}_t\right)\Big\},
    \end{align}
    where $d_H\left(p^{(i)}_t,{p}_{t+1}\right) = \min_{p^{(j)}_{t+1}\in{p}_{t+1}}d\left(p^{(i)}_t, p^{(j)}_{t+1}\right)$ quantifies the distance from a point $p^{(i)}_t$ to the point set ${p}_{t+1}$, and we adopt the Euclidean distance to implement $d(\cdot, \cdot)$. Last, the overall temporal consistency of an animation $x$  is computed by averaging frame-wise temporal consistency scores:
    \begin{align}
        d_H(x) = \frac{1}{N-1}\sum_{t = 0}^{N-2}d_H({p}_t, {p}_{t+1}),
    \end{align}
    where higher values indicate poorer temporal consistency.
    \item \textit{Geometric Deviation}.
    We measure the geometric deviation in an animation by analyzing its local discrete curvature changes at control points that define the object. Specifically, let $v^{(i\rightarrow i-1)}_t$ and $v^{(i+1 \rightarrow {i})}_t$ be the vectors of the $i$-th control point at the $t$-th frame, pointing to its two adjacent points. The curvature $\kappa^{(i)}_t$ of at this control point is 
    \begin{align}
        \kappa^{(i)}_t = \frac{\theta^{(i)}_t}{\left\Vert v^{(i\rightarrow i-1)}_t\right\Vert_2 + \left\Vert v^{(i+1 \rightarrow {i})}_t\right\Vert_2},
    \end{align}
    where $\theta^{(i)}_t$ is the angle between $v^{(i\rightarrow i-1)}_t$ and $v^{(i+1 \rightarrow {i})}_t$. The geometric deviation of $x$ is then computed by the mean absolute differences in curvature between the initial and animated frames:
    \begin{align}
        \mathrm{GD}(x) = \frac{1}{(N-1)L}\sum_{t=1}^{N-1}\sum_{i=0}^{L-1}\left\vert\kappa_0^{(i)} -\kappa_t^{(i)} \right\vert,
    \end{align}
    where $L$ is the total number of control points defining the object in the initial frame. Larger values correspond to more severe geometric distortions.
    
\end{enumerate}

\subsection{Main Results}
\label{experiment_comparison}

We compare \sysName~with two alternatives: (1) Gal23~\citep{gal2023breathing} for vector sketch animation and (2) state-of-the-art T2V models.

\begin{table}[b]
\centering
\small
\begin{tabular}{lcc}
    \toprule
    Method & \begin{tabular}[c]{@{}c@{}}Visual Identity\\ Preservation\\ (CLIP Score $\uparrow$) \end{tabular}  & \begin{tabular}[c]{@{}c@{}}Text-Video\\ Alignment\\ (X-CLIP Score $\uparrow$) \end{tabular} \\
    \noalign{\vskip 1mm}
    \toprule
    Gal23 & $0.8379$ & $0.1879$ \\
    ModelScope & $0.8618$ & $0.2021$ \\
    DynamiCrafter & $0.8008$ & $0.1741$ \\
    I2VGen-XL & $0.8813$ & $0.1999$ \\
    VideoCrafter2 & $0.8403$ & $0.1992$ \\
    ToonCrafter & $0.9292$ & $0.2003$ \\
    \noalign{\vskip 0.5mm}
    \hline
    \noalign{\vskip 0.5mm}
    \sysName~(Ours) & $\mathbf{0.9414}$ & $\mathbf{0.2071}$ \\
    \bottomrule
\end{tabular}
\caption{
Quantitative results of \sysName~in terms of bitmap metrics against Gal23 and T2V models.
}
\label{tab:comparison}
\end{table}

\noindent{\textbf{\sysName~versus Gal23}}.
Both our \sysName~and Gal23 aim to leverage implicit motion priors in pretrained T2V models with the video SDS loss. However, they differ substantially in two key aspects. First, Gal23 does not enforce object rigidity, leading to noticeable geometric distortions in the overall appearance across frames, along with color cast distortions, as shown in Fig.~\ref{fig:result_anisketch}.
While minor geometric distortions may be acceptable for abstract sketches, they are highly detrimental for clipart animation, which expects precise geometry reproduction.
In contrast, \sysName~explicitly enforces rigidity by incorporating and differentiating through ARAP shape deformation during animation. Second, 
Gal23 directly predicts future control points in a sketch, whereas \sysName~keeps track of a significantly smaller set of keypoints (\eg, $13$ versus $1,067$ for the fencer example in Fig.~\ref{fig:pipeline}) using cubic B\'{e}zier parameterization. Through end-to-end optimization guided by the combined video SDS loss and skeleton fidelity loss, \sysName~produces identity-preserved, text-aligned, motion-consistent, and cartoon-like clipart animations, which is further evidenced by the highest CLIP and X-CLIP scores in Table~\ref{tab:comparison}.

\begin{figure*}[t]
    \centering
    \includegraphics[width=0.95\linewidth]{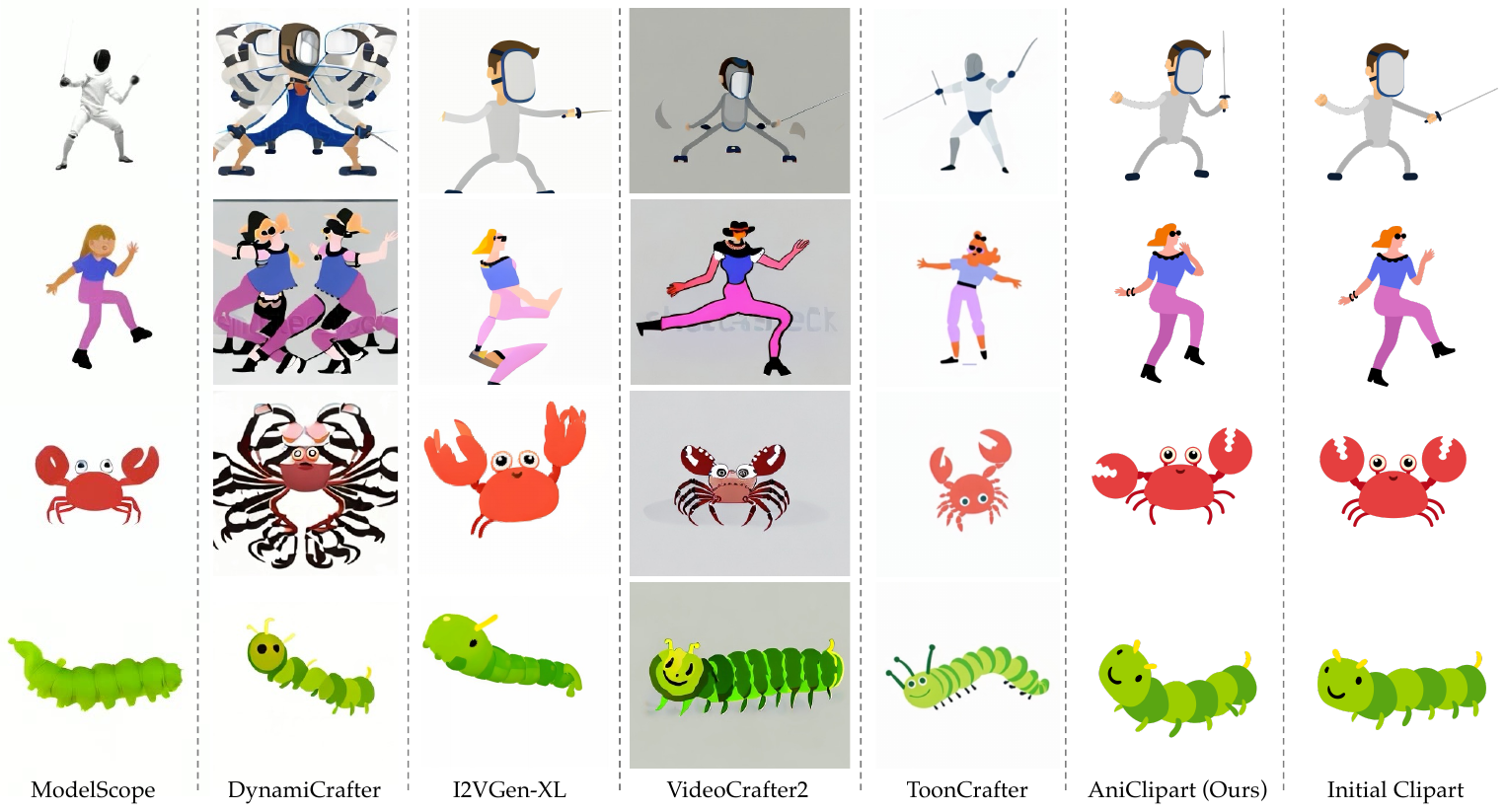}
    \caption{\textbf{\sysName~versus T2V Models}.
    T2V models focus primarily on preserving the semantics of initial clipart images but often neglect finer details, which result in poor-quality outputs with a lack of identity preservation. Furthermore, they frequently yield animations with minimal motion, leading to weak text-video alignment. In contrast, \sysName~addresses these shortcomings, delivering high-quality clipart animations.
    }
    \label{fig:result_i2v}
\end{figure*}

\begin{figure}[t]
    \centering
    \includegraphics[width=1.0\linewidth]{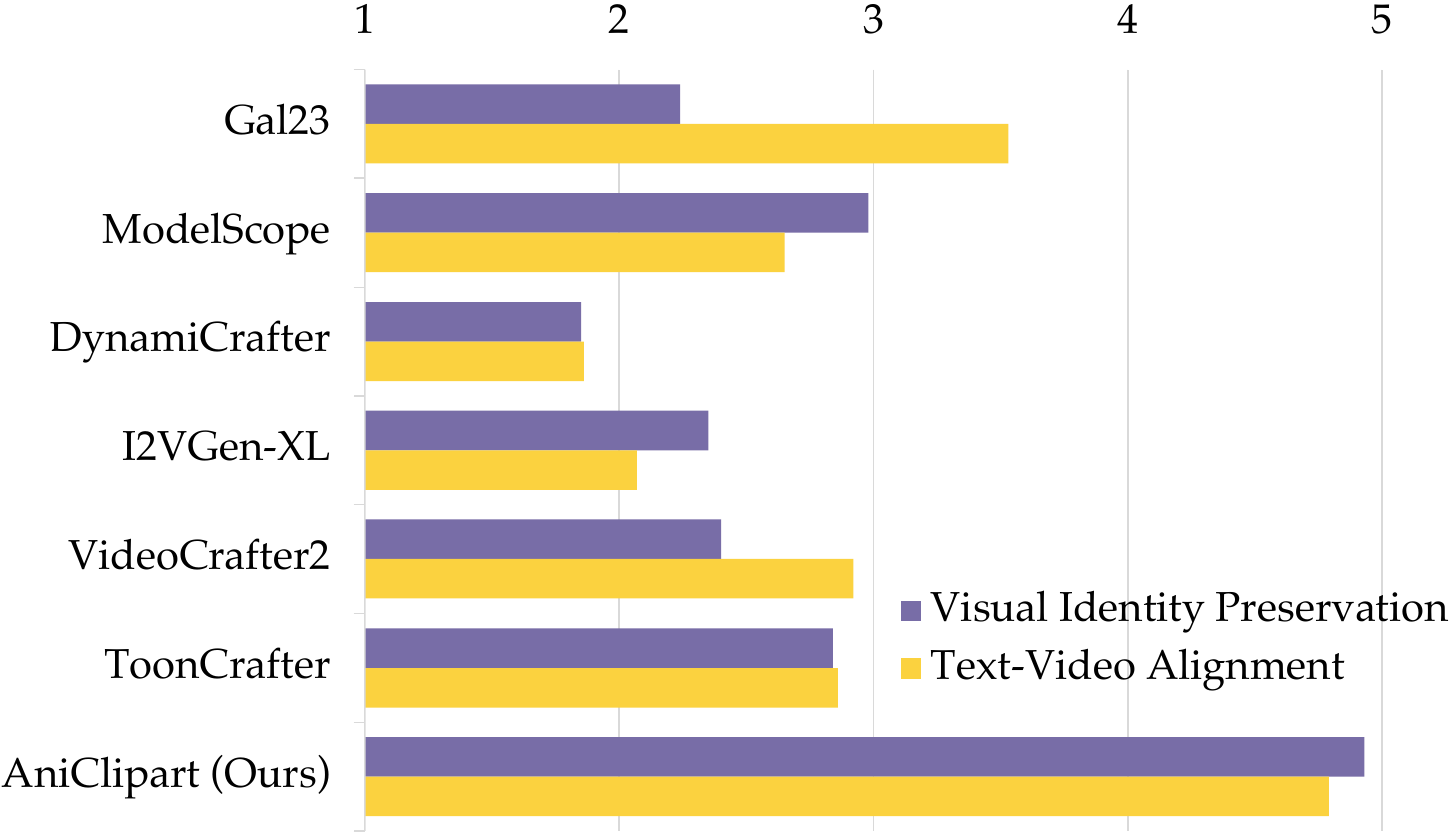}
    \caption{Quantitative results of our subjective user study.}
    \label{fig:user_study}
\end{figure}

\noindent{\textbf{\sysName~versus T2V Models}}.
In our experiments, we rasterize vector clipart images, and feed them to T2V models, including ModelScope~\citep{wang2023modelscope}, DynamiCrafter~\citep{xing2023dynamicrafter}, I2VGen-XL~\citep{zhang2023i2vgen}, VideoCrafter2~\citep{chen2024videocrafter2}, and ToonCrafter~\citep{xing2024tooncrafter} for clipart animation.
In particular, ModelScope cannot accept an image as an additional condition alongside the text prompt. 
To address this, we adopt the approach proposed in SDEdit~\citep{meng2021sdedit} by introducing random noise into the image and using the noise-injected image as input to ModelScope.
Empirically, we set the blending ratio to $0.84$. Increasing this ratio results in larger motion but compromises the preservation of visual identity, while decreasing the ratio better retains visual identity but creates minimal motion, leading to nearly static objects.
ToonCrafter, fine-tuned on an anime dataset with video interpolation capabilities,  requires both the first and last frames as input. For the starting frame, we use the original clipart image. To generate the last frame automatically, we employ IP-Adapter~\citep{ye2023ip}, which creates variations of the input object with consistent content and style.

Fig.~\ref{fig:result_i2v} visually compares \sysName~with T2V models on representative animated frames.
Two primary shortcomings of T2V models are evident. First, T2V models often generate visually annoying spatial artifacts, such as distorted shapes and blurred details, which compromise the visual identity of the initial objects. Second, motion generated by T2V models is generally less dynamic than that of \sysName.
For example, while ModelScope strikes a reasonable balance between identity preservation and motion generation by carefully tuning the blending ratio, it occasionally fails to induce any movement for the objects, resulting in inferior text-video alignment (see Table \ref{tab:comparison}).
DynamiCrafter tends to add unwanted shapes and textures to initial clipart images without introducing visible and semantically meaningful motion.
I2VGen-XL can produce moderate motion; however, it frequently distorts the object appearance (\eg, the dancing woman).
VideoCrafter2 occasionally produces reasonable results, but in many cases, it fails to animate the objects.
ToonCrafter, though adept at creating animations with large motion and consistent style, faces challenges in preserving object identity due to the inherent limitations of IP-Adapter that provides identity-compromised last frames. Additionally, the interpolated animations by ToonCrafter are mainly characterized by linear motion, in contrast to complex, vivid, and cartoon-like motion by \sysName. In summary, ``zero-shot prompting'' of T2V models for clipart animation is generally impractical, highlighting the need for specialized models like \sysName~for this particular task.

\noindent{\textbf{Subjective User Study}}.
To quantitatively assess the improvements in animation quality by \sysName, we conducted a formal subjective user study, which includes two tasks aimed at evaluating (1)  visual identity preservation and (2)  text-video alignment. We selected $30$ static clipart images, each animated by seven different methods, including Gal23, ModelScope, DynamiCrafter, I2VGen-XL, VideoCrafter2, ToonCrafter, and \sysName.
For the first task, participants were shown seven animations of the same object along with the initial clipart image at one time, and asked to rate the quality of each animation, focusing primarily on visual identity preservation. 
For the second task, the input text prompt was provided alongside the initial clipart image, and participants were asked to rate how well each animation aligned with the given text prompt. Ratings were collected using a five-point Likert scale, with $1$ representing ``Strongly Disagree'' and $5$ representing ``Strongly Agree''.
The subjective user study was implemented as an online questionnaire, where a total of $42$ participants were invited. To minimize fatigue effects, participants were allowed to take breaks at any time.

Fig.~\ref{fig:user_study} shows the subjective evaluation results, averaged across $30$ clipart examples. It is clear that the proposed \sysName~significantly outperforms the competing methods in terms of both visual identity preservation and text-video alignment. The statistical significance of the performance improvements has been confirmed through a one-way ANalysis Of VAriance (ANOVA) test.

\begin{table*}[t]
\centering
\small
\begin{tabular}{lccc}
    \toprule
    \multicolumn{1}{c}{\sysName~Variant} & Motion Vibrancy $\uparrow$ & Temporal Consistency $\downarrow$ & Geometric Deviation $\downarrow$ \\
    \noalign{\vskip 1mm}
    \toprule
    \noalign{\vskip 1mm}
    ARAP Deformation $\rightarrow$ Linear Blend Skinning & $17.13$ & $5.4631$ & $60.56$ \\
    w/o B\'{e}zier Trajectory Parameterization & $\mathbf{102.66}$ & $13.9341$ & $65.22$ \\
    First-order B\'{e}zier Trajectory Parameterization & $8.96$ & $\mathbf{4.7950}$ & $\mathbf{42.42}$ \\
    Video SDS Loss $\rightarrow$ X-CLIP Loss & $8.16$ & $6.1486$ & $108.81$ \\
    Video SDS Loss $\rightarrow$ Image SDS Loss & $7.57$ & $5.6211$ & $59.07$ \\
    w/o Skeleton Fidelity Loss & $16.18$ & $8.3768$ & $265.92$ \\
    \noalign{\vskip 0.5mm}
    \hline
    \noalign{\vskip 0.5mm}
    Default & $20.87$ & $8.5115$ & $50.98$ \\
    \bottomrule
\end{tabular}
\caption{
Quantitative results of \sysName~variants.
}
\label{tab:ablation}
\end{table*}

\begin{figure*}[t]
    \centering
    \includegraphics[width=0.82\linewidth]{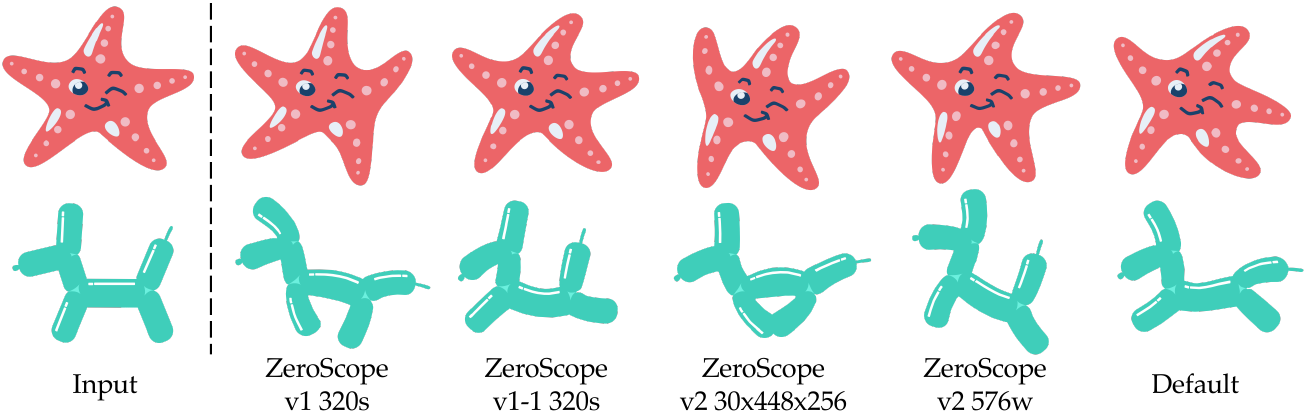}
    \caption{\textbf{Effects of Different T2V Backbones}.
    We display the last frames from animations generated by various T2V models, highlighting noticeable motion variations. In the accompanying annotations, ``s'' indicates that inputs are square-shaped videos, while ``w'' refers to the frame width.
    }
    \label{fig:video_backbone}
\end{figure*}

\subsection{Ablation Studies} \label{experiment_ablation}

In this subsection, we present a series of ablation studies to justify the design choices of \sysName, with the quantitative results listed in Table~\ref{tab:ablation} and qualitative results shown in Fig. \ref{fig:ablation}.

\noindent{\textbf{ARAP Shape Deformation}}.
To analyze the effectiveness of ARAP shape deformation, we replace it with linear blend skinning, another widely-used algorithm for shape deformation. It predicts future triangle vertices in a mesh as a linear weighted sum of the updated keypoint positions, where the bounded biharmonic weights~\citep{jacobson2011bounded} are used. However, linear blend skinning frequently produces unrealistic deformation, such as the squeezed shoulders of the dancing man and the crippled legs of the dog in Fig. \ref{fig:ablation}. Such geometric distortions are also highlighted by the vector metrics in Table~\ref{tab:ablation}. The results of the model variant without any structured 2D deformation are close to those by Gal23 (see Fig. \ref{fig:result_anisketch}), which is visually inferior to the full \sysName.

\noindent{\textbf{B\'{e}zier Trajectory Parameterization}}.
We parameterize the keypoint motion trajectories using cubic B\'{e}zier curves to achieve smooth, natural animations. As a comparison, we take a non-parametric approach similar to Gal23, in which we directly predict future keypoints without using trajectory parameterization. While this variant can produce animations aligned with text descriptions, it fails to maintain temporal consistency, causing noticeable flickering artifacts (see the ``w/o B\'{e}zier Trajectory Parameterization'' column in Fig. \ref{fig:ablation}). This deficiency explains the highest motion vibrancy score, coupled with the poorest temporal consistency score, as shown in Table \ref{tab:ablation}.
We also visualize in Fig.~\ref{fig:ablation} the resulting jittery pseudo-trajectories by connecting the predicted keypoints linearly over time.
Next, we simplify B\'{e}zier trajectory parameterization to first order, which produces overly simplistic linear motion. While this yields the greatest temporal consistency and the least geometric deviation in Table \ref{tab:ablation}, the limited expressiveness of linear motion makes it less capable of adhering to the nuanced and semantically rich text conditions (see also Fig.~\ref{fig:complex_path}). For a more direct comparison, kindly refer to the side-by-side video examples at \url{https://aniclipart.github.io/}.

\noindent{\textbf{Loss Functions}}.
The video SDS loss in Eq.~\eqref{eq:lSDS} is crucial for generating text-aligned motion trajectories. In contrast,
replacing it with another motion-aware loss function---the X-CLIP loss~\citep{ni2022expanding}---does not produce reasonable text-aligned motion, leading instead to abnormal movements with severe distortions.
Additionally, substituting the video SDS loss with an image-based counterpart, implemented by the Stable Diffusion v1.5~\citep{rombach2022high}, further degrades the animations’ text relevance.
Moreover, omitting the skeleton fidelity loss in Eq.~\eqref{eq:fidelity} compromises anatomical accuracy, resulting in unnatural bone proportions.

\noindent{\textbf{Video Models}}.
In our experiments, we employ the publicly available ModelScope~\citep{wang2023modelscope} as our default T2V backbone.
We also explore other T2V models, such as ZeroScope\footnote{\url{https://huggingface.co/cerspense}}, which has been fine-tuned on videos across diverse resolutions and framerates. Fig.~\ref{fig:video_backbone} presents the last frames of animations produced by different T2V  models, revealing variations in motion patterns, yet all remain faithful to the given text prompt. Consequently, \sysName~can seamlessly integrate with different T2V diffusion models, and thus capitalize on the latest advances in T2V generation.

\subsection{Extensions} \label{extension}
We suggest two extended implementations for \sysName~that broaden its range and improve the overall quality of the generated animations.

\begin{figure}
    \centering
    \includegraphics[width=0.95\linewidth]{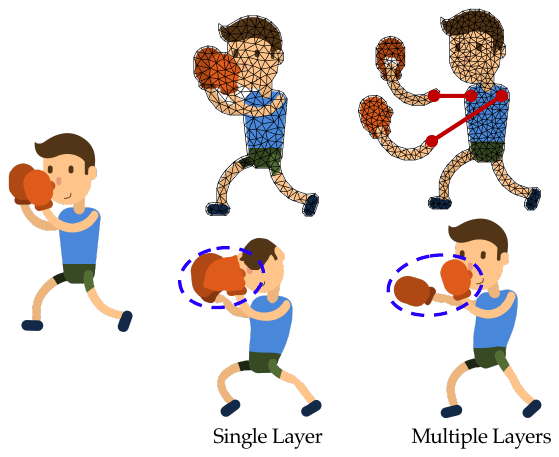}
    \caption{
    \textbf{Layered Animation}.
    The multi-layer variant of \sysName~enables clipart animation involving topological changes, and significantly reduces geometric distortions, which are otherwise clearly observed in the result by the default single-layer \sysName.
    }
    \label{fig:layered_mesh}
\end{figure}

\noindent{\textbf{Layered Animation}}.
In clipart preprocessing, we currently create a single triangle mesh for shape deformation (see Sec. \ref{method_data_preperation}).
However, this implementation is less effective in handling topological changes (\eg, caused by self-occlusions) in animation. For instance, in Fig.~\ref{fig:layered_mesh}, when the character's right hand overlaps with the left,  a single mesh cannot effectively separate the hands to achieve a desired boxing animation.
To address these issues, we recommend a multi-layer animation pipeline by making the following modifications.
\begin{enumerate}
    \item Group paths in vector clipart into semantically meaningful layers (\eg, the body). This is done manually because no automated algorithm currently exists that can reliably identify meaningful, movable parts in SVG.
    \item The method for detecting keypoints remains unchanged. However, since each layer corresponds to a distinct object part, we just determine the boundary for each layer, and locate the keypoints within that boundary. Doing so ensures that each layer is updated accurately based on its own keypoints.
    \item Similarly, we build a triangle mesh for each layer, which is deformed separately and combined to produce the next clipart frame.
\end{enumerate}
Fig.~\ref{fig:layered_mesh} shows the visual improvements by the multi-layer variant of \sysName. Notably, as we operate on vector clipart, there is no need to reconstruct previously occluded parts during animation---each SVG path is already complete.

\begin{figure}
    \centering
    \includegraphics[width=0.95\linewidth]{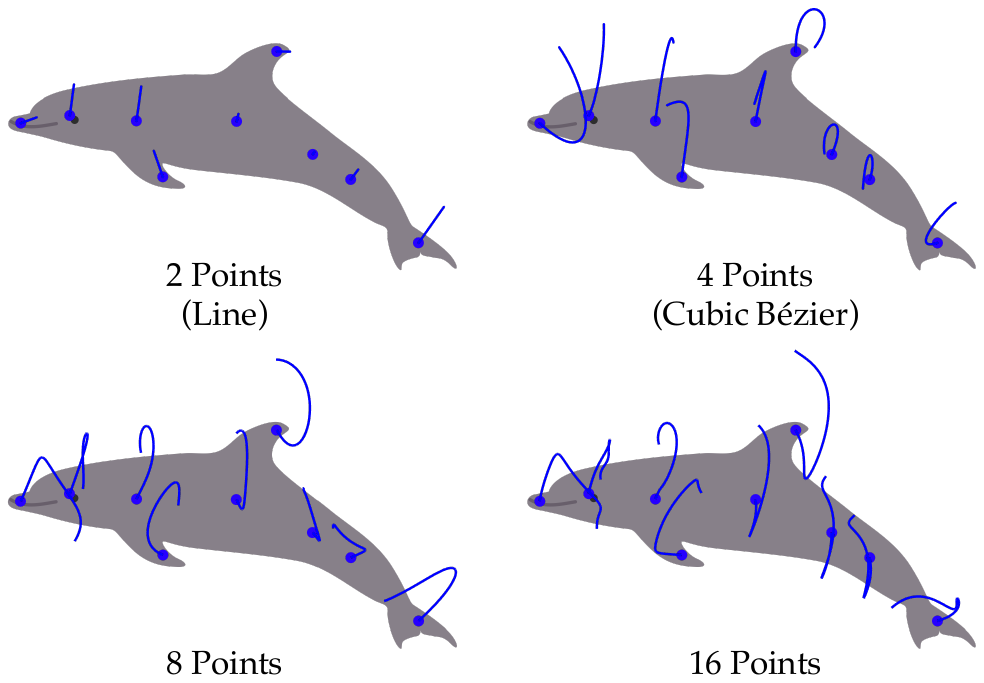}
    \caption{High-order B\'{e}zier trajectories allow for more complex, precise motion synthesis by increasing control points.}
    \label{fig:complex_path}
\end{figure}

\noindent{\textbf{Higher-Order B\'{e}zier Trajectories}}.
In our experiments, we adopt the cubic B\'{e}zier trajectories (each defined by four control points) to regularize keypoint motion for clipart animation.
Our approach works naturally with higher-order B\'{e}zier trajectories by adding control points to enable more complex, precise motion synthesis. Fig.~\ref{fig:complex_path} shows such a visual example.  In response to finer animation details, more frames need to be sampled. We find that memory consumption scales linearly with the number of frames, while the time required per animation remains relatively the same.
After examining various animations, we find that cubic B\'{e}zier trajectories strike an excellent balance between ease of use and visual quality. In practice, this trade-off should be left to designers, allowing them to determine the most suitable settings for their particular scenarios.

    % --------------------------------------------------------

    %--------------------------------------------------------
    \section{Conclusion and Discussion} \label{conclusion}

In this work, we have introduced \sysName~for text-driven clipart animation. \sysName~first defines keypoints on the input clipart image, then employs B\'{e}zier curves to parameterize motion trajectories. Crucially, it derives motion priors from pretrained T2V diffusion models using the video SDS loss, without resorting to any specialized training data. Additionally, we incorporate a skeleton fidelity loss to encourage motion coherence among keypoints, which then drives clipart animation through ARAP shape deformation. Comprehensive experiments confirm the effectiveness of \sysName~in synthesizing high-quality clipart animations.

\begin{figure}[t]
    \centering
    \includegraphics[width=0.9\linewidth]{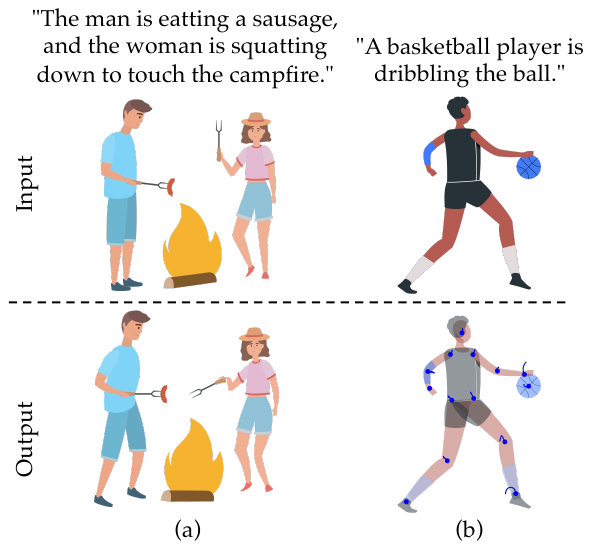}
    \caption{\textbf{Limitations}. \sysName~experiences a performance decline when handling complex clipart with multiple objects, due in part to the inaccuracy of the video SDS loss.}
    \label{fig:limitation}
\end{figure}

\noindent\textbf{Limitations}. \sysName~encounters difficulties when handling complex clipart featuring multiple objects.
For instance, in Fig.~\ref{fig:limitation} (a), which shows two characters at a picnic, the synthesized motion often appears unnatural, and the resulting animation may fail to align with the input text prompt.
A similar issue is observed in Fig.~\ref{fig:limitation} (b), even when we place the man and the basketball in separate layers. Although our goal is to animate the man dribbling the ball, the basketball never touches the ground. This performance degradation largely stems from the inability of the video SDS loss to understand the physical laws of motion, resulting in text-misaligned animations with visible artifacts. We expect a more advanced T2V model could help overcome these limitations and improve the quality of animations.

\noindent\textbf{Future Work}.
In the future, we aim to further automate our clipart animation pipeline to better support designers.
Currently, our keypoint detection relies on a hybrid approach that combines both template-based and template-free methods. To streamline this step, we plan to develop a keypoint detection method tailored for vector clipart by learning from clipart-keypoint pairs.
We also intend to reduce the manual effort involved in layered animation. This may be achieved by multimodal large language models, such as GPT-o1, that automatically segment vector clipart into semantically meaningful, movable parts. Looking ahead, we seek to broaden the capabilities of \sysName~to tackle more complex tasks. Integrating 2.5D cartoon models~\citep{rivers20102} would enable the simulation of 3D rotation effects within 2D clipart, adding depth to the resulting animations. Finally, we plan to investigate techniques for better animating complex scenes featuring multiple objects, thus extending the real-world applicability of \sysName.

% IJCV specific
\noindent\textbf{Data Availability}.
We have included data ($30$ SVG clipart images) as electronic supplementary material.

\begin{acknowledgement}
The work described in this paper was fully supported by a GRF grant from the Research Grants Council (RGC) of the Hong Kong Special Administrative Region, China [Project No. CityU 11216122].
\end{acknowledgement}
    %--------------------------------------------------------
    
    \bibliographystyle{spbasic}      % basic style, author-year citations 
    \bibliography{main.bib}

    \end{document}